\documentclass[11pt]{article}
\usepackage{amsfonts}
\usepackage{mathrsfs}
\usepackage{amsmath,amssymb}
\usepackage{mathtools}

\DeclareMathOperator{\EX}{\mathbb{E}}

\usepackage[latin1]{inputenc}
\usepackage{amsmath,amssymb}

\usepackage{amsthm}
\usepackage{latexsym}
\usepackage{epstopdf}
\usepackage{geometry}
\usepackage{hyperref}
\usepackage{pict2e,picture}
\usepackage{bigints}
\usepackage[thinlines]{easytable}
\usepackage{enumerate}
\usepackage{array}
\usepackage{fix-cm}
\usepackage{tikz}
\usetikzlibrary{matrix}

\newcolumntype{C}[1]{>{\centering\arraybackslash}m{#1}}

\newcolumntype{R}[1]{>{\raggedleft\arraybackslash}m{#1}}

\usepackage{lipsum}

\usepackage[symbol]{footmisc}

\usepackage[active]{srcltx}

\usepackage{graphicx}

\usepackage{epstopdf}

\usepackage{enumitem}

\usepackage[T1]{fontenc}

\usepackage{amsmath}

\usepackage{footnote}

\usepackage{footmisc}

\makesavenoteenv{tabular}

\makesavenoteenv{table}

\usepackage{bbm}

\DeclareMathOperator*{\argmin}{arg\,min}

\newtheorem {theorem}{Theorem}[section]

\newtheorem{remark}{Remark}[section]

\renewcommand\footnotemark{}

\usepackage{dsfont}

\date{\vspace{-5ex}}

\begin{document}

\title{Function approximation and nonparametric regression with binary and ternary ReLU networks}

\maketitle

\begin{center}

\bigskip Aleksandr Beknazaryan \footnote{abeknazaryan@yahoo.com}

\bigskip

\end{center}

\begin{abstract}

\vskip.2cm
We show that deep binary ReLU networks and deep sparse ternary ReLU networks can approximate $\beta$-H\"older functions on $[0,1]^d$. We also show that deep sparse ternary ReLU networks can achieve, up to a logarithmic factor, the minimax rate of prediction of $\beta$-smooth regression function.
\vskip.2cm \noindent 


\vskip.2cm

\end{abstract}

\section{Introduction}
Deep neural networks have proven to be elegant and effective computational tools for dealing with problems of classification, clustering, pattern recognition and prediction.
Improvement of the accuracy of performance of DNN models usually results in significant growth of these models making them computationally expensive and memory intensive. 
Realizations of deep models on low-capacity devices with limited computational power can be done by restricting the possible values of network weights. Popular types of such networks are binary and ternary weighted neural networks with weights that usually belong to the sets $\{\pm1\}$ and $\{0, \pm1\}$, respectively. Due to extreme memory reduction, high computational speed and energy efficiency, binary and ternary networks have attracted a lot of interest in recent scientific literature.
In particular, various training algorithms as well as applications of such networks in image classification, natural language processing, speech recognition and molecular programming were given in the works \cite{A}, \cite{BZ}, \cite{C} --  \cite{LZSC}, \cite{SL} -- \cite{X} and \cite{YA}. Also, a lower bound for the number of linear regions of ternary ReLU regression neural networks was recently derived in \cite{N}.

Our goal is to describe the expressive power of binary and ternary ReLU networks. The expressivity of deep ReLU networks has been well studied in the recent literature with a specific focus on achieving high approximation and prediction accuracy by considering sparse networks and, thus, reducing the parameter complexity. In particular, in \cite{Y} it was shown that on $[0,1]^d$ ReLU networks of depth $O(\ln(1/\varepsilon))$ with $O(\varepsilon^{-d/n}\ln(1/\varepsilon))$ nonzero weights can $\varepsilon$-approximate functions from the unit ball of the Sobolev space $\mathcal{W}^{n, \infty}.$ Also, in \cite{SH} it was shown that functions from the H\"older ball $\mathcal{C}^\beta_d(K)$ can be $\varepsilon$-approximated by deep sparse ReLU networks of depth $O(\ln(1/\varepsilon))$ with $O(\varepsilon^{-d/\beta}\ln(1/\varepsilon))$ nonzero weights all which are in the interval $[-1,1]$. The boundedness of network weights then leads to minimax optimal rates of prediction (up to a logarithmic factor) of $\beta$-smooth regression functions by deep sparse ReLU networks. Identical rates of approximation and regression estimation by deep sparse ReLU networks with weights $\{0,\pm1/2, \pm 1, 2\}$ were derived in \cite{B}. 

Here we show that the approximations constructed in \cite{B} can be realized by binary and ternary ReLU networks. 
In particular, we show that deep sparse ternary ReLU networks can achieve, up to a logarithmic factor, the minimax rate of prediction of $\beta$-smooth regression function.

\textit{Notation.} For a vector $\textbf{v}\in\mathbb{R}^d$ and a function $f$ on $[0,1]^d$, the notations $|\textbf{v}|_\infty$ and $\|f\|_{\infty}$ denote, respectively, the $l_\infty$ norm of $\textbf{v}$ and the sup norm of $f$ on $[0,1]^d$, $d\in\mathbb{N}$. Also, the centered dot $\cdot$ denotes the usual matrix multiplication.

\section{Function approximation with binary and ternary ReLU networks}

Consider the class $$\mathcal{F}(L,\textbf{p}):=\bigg\{f:[0,1]^d\to\mathbb{R}\; |\; f(\textbf{x})=W_L\cdot\sigma\circ W_{L-1}\cdot\sigma\circ...\cdot\sigma\circ W_0\cdot(1, \textbf{x})\bigg\}$$
of feedforward ReLU networks of depth $L$, where in each hidden layer the ReLU activation function $\sigma(x)=\max\{0,x\}$ acts coordinate-wise on the input vectors: $\sigma\circ(z_1,...,z_r)=(\sigma(z_1),...,\sigma(z_r)),$ and the coordinate $1$ added to the input $\textbf{x}$ allows to omit the shift vectors. The \textit{width vector} $\textbf{p}=(p_0, p_1,...,p_{L+1}),$ with $p_0=d+1$ and $p_{L+1}=1,$ contains the sizes of the \textit{weight matrices} $W_i\in\mathbb{R}^{p_{i+1}\times p_{i}}$, $i=0,...,L,$ and the entries of the weight matrices are called the \textit{weights} of the network $f$. For a set $\mathcal{A}\subset\mathbb{R}$ let
$$\mathcal{F}_\mathcal{A}(L,\textbf{p}):=\big\{f\in\mathcal{F}(L,\textbf{p})\;| \textrm{ all the weights of } f \textrm{ are in } \mathcal{A} \big\}$$ and for $s\in\mathbb{N}$ let $\mathcal{F}_\mathcal{A}(L,\textbf{p}, s)$ be the subset of $\mathcal{F}_\mathcal{A}(L,\textbf{p})$ consisting of networks with at most $s$ nonzero weights.
Approximations of $\beta$-H\"older functions belonging to the ball 
$$\mathcal{C}^\beta_d(K):=\bigg\{f:[0,1]^d\to\mathbb{R}: \sum\limits_{0\leq|\boldsymbol{\alpha}|<\beta}\|\partial^{\boldsymbol{\alpha}}f\|_{\infty}+\sum\limits_{|\boldsymbol{\alpha}|=\lfloor\beta\rfloor}\sup\limits_{\substack{\textbf{x},\textbf{y}\in[0,1]^d \\ \textbf{x}\neq \textbf{y}}}\frac{|\partial^{\boldsymbol{\alpha}}f(\textbf{x})-\partial^{\boldsymbol{\alpha}}f(\textbf{y})|}{|\textbf{x}-\textbf{y}|_\infty^{\beta-\lfloor\beta\rfloor}}\leq K\bigg\}$$ with sparse networks from $\mathcal{F}_{[-1,1]}(L,\textbf{p}, s)$ are established in \cite{SH}. Using that result and substituting the interval $[-1,1]$ by the finite set of weights $\{0,\pm \frac{1}{2}, \pm 1, 2\}$, the following theorem is given in \cite{B}:
\begin{theorem}\label{prev}
For any function $f_0\in\mathcal{C}^\beta_d(K)$ and any integers $m\geq 1$ and $N\geq(\beta+1)^d\lor(K+1)e^d,$ there exists a ReLU network $f\in\mathcal{F}_\mathcal{A}(L,\normalfont\textbf{p}, s)$ with $\mathcal{A}=\{0,\pm \frac{1}{2}, \pm 1, 2\},$
$$L=16+2(m+5)(1+\lceil\log_2(d\lor\beta)\rceil)+8\log_2(N^{\beta+d}Ke^d),$$
$$|\normalfont{\textbf{p}}|_\infty=\bigg[2\bigg(1+d+(2\beta)^dN+2\log_2(N^{\beta+d}Ke^d)\bigg)\bigg]\lor\bigg[2^d6(d+\lceil\beta\rceil)N\bigg],$$
having at most $s=141(d+\beta+1)^{3+d}L|\normalfont{\textbf{p}}|_\infty$
nonzero weights, such that
$$\|f-f_0\|_{\infty}\leq C(N2^{-m}+ N^{-\frac{\beta}{d}}),$$
where $C=C(\beta, d, K)$ is some constant.
\end{theorem}

In the following theorem we show that the above approximation can be done using deep sparse ReLU networks with weights ${\{0,\pm1\}}$ and with binary ReLU networks with weights ${\{\pm1\}}$.
\begin{theorem}\label{main}
For any function $f_0\in\mathcal{C}^\beta_d(K)$ and any integers $m\geq 1$ and $N\geq(\beta+1)^d\lor(K+1)e^d,$ there exist networks $f_1\in1/2^{L+3}\cdot\mathcal{F}_{\{0,\pm1\}}(L+2,\normalfont\textbf{p}', s')$ and $f_2\in1/4^{L+6}\cdot\mathcal{F}_{\{\pm1\}}(L+5,\normalfont\textbf{p}'')$ with
$$|\normalfont{\textbf{p}}^\prime|_\infty=4|\normalfont{\textbf{p}}|_\infty, \;\;\; |\normalfont{\textbf{p}}^{\prime\prime}|_\infty=32|\normalfont{\textbf{p}}|_\infty\;\;\; \textrm{and} \;\;\; s^\prime\leq16s+20(d+1),$$
such that $$\|f_i-f_0\|_{\infty}\leq C(N2^{-m}+ N^{-\frac{\beta}{d}}),\;\;\; i=1,2,$$
where $L, \normalfont\textbf{p}$ and $s$ are the same as in Theorem \ref{prev} and $C=C(\beta, d, K)$ is some constant.\end{theorem}
\begin{remark} Note that since the ReLU function $\sigma$ is positive homogeneous, that is, $\sigma(ax)=a\sigma(x)$ for any $a>0, x\in\mathbb{R}$, then, to avoid the scaling factors $1/2^{L+3}$ and $1/4^{L+6}$, Theorem \ref{main} can also be restated for ternary and binary ReLU networks with weights   $\{0, \pm1/2\}$ and $\{\pm1/4\}$. Alternatively, the scaling factors can be omitted by considering networks with weights $\{0, \pm1\}$ and $\{\pm1\}$ and with activation functions $\sigma/2$ and $\sigma/4$, respectively. 
\end{remark}

\section{Nonparametric regression with sparse ternary ReLU networks}
In this section we use the result of Theorem \ref{main} to show that an unknown regression function $f_0:[0,1]^d\to[-F, F], f_0\in\mathcal{C}^\beta_d(K),$ $F\geq\max(K,1),$ can be recovered using deep sparse ternary ReLU networks with weights ${\{0,\pm1\}}$. The network architecture and the number of nonzero weights in the networks are determined by the sample size $n$ of observed iid pairs $(\textbf{X}_i, Y_i)$ following a nonparametric regression model 
\begin{equation}\label{model}
    Y_i=f_0(\textbf{X}_i)+\epsilon_i, \;\;\; i=1,...,n,
\end{equation}
where the standard normal noise variables $\epsilon_i$ are assumed to be independent of $\textbf{X}_i$. More precisely, for $m_n:=\lceil \log_2n\rceil$ and $N_n:=\lceil (\beta+1)^d\lor(K+1)e^d\lor n^{\frac{d}{2\beta+d}}\rceil,$ let $\mathcal{{F}}_{\{0,\pm1\}}({L}_n,\normalfont{\textbf{p}}_n,{s}_n)$ be the space of ReLU networks with weights $\{0,\pm1\}$ and with  
\begin{equation}\label{ln}
    \textit{the depth } L_n:= 16+2(m_n+5)(1+\lceil\log_2(d\lor\beta)\rceil)+8\log_2(N_n^{\beta+d}Ke^d)\sim\log_2n,
\end{equation}
\begin{equation}\label{pn}
\textit{the width }|\normalfont{\textbf{p}}_n|_\infty:=4\Bigg(\bigg[2\bigg(1+d+(2\beta)^dN_n+2\log_2(N_n^{\beta+d}Ke^d)\bigg)\bigg]\lor\bigg[2^d6(d+\lceil\beta\rceil)N_n\bigg]\Bigg)\sim n^{\frac{d}{2\beta+d}}\end{equation}
and with 
\begin{equation}\label{sn}
\textit{the number of nonzero weights }s_n\leq 2260(d+\beta+1)^{3+d}L_n|\normalfont{\textbf{p}_n}|_\infty\sim n^{\frac{d}{2\beta+d}}\log_2n.\end{equation} 

Then, for $\mathcal{F}_n:=1/2^{L_n+3}\cdot\mathcal{{F}}_{\{0,\pm1\}}({L}_n+2,\normalfont{\textbf{p}}_n,{s}_n)$ we have the following 
\begin{theorem}\label{reg}
    The prediction error 
    $$R(\hat{f}_n,f_0)=\EX_{f_0}[(\hat{f}(\textbf{X})-f_0(\textbf{X}))^2]$$
    of the empirical risk minimizer  
    \begin{equation}\label{erm}
        \hat{f}_n\in\argmin_{f\in\mathcal{F}_n}\sum_{i=1}^{n}(Y_i-f(\textbf{X}_i))^2
    \end{equation} 
    over the space $\mathcal{F}_n$ is bounded by
    \begin{equation}\label{err}
R(\hat{f}_n,f_0)\leq Cn^{\frac{-2\beta}{2\beta+d}}\log_2^2n,
\end{equation} where $C=C(\beta, d, F)$ is some constant and $\textbf{X}\stackrel{\mathcal{D}}{=}\textbf{X}_1$ is independent of the sample $(\textbf{X}_i, Y_i)$.
\end{theorem}

 Note that the upper bound given in \eqref{err} coincides, up to a logarithmic factor, with the minimax rate $n^{\frac{-2\beta}{2\beta+d}}$ of the prediction error of $\beta$-smooth functions.

\section{Proofs}
\textit{Proof of Theorem \ref{main}}
The positive homogeneity of the ReLU function $\sigma$ implies that $$1/2^{L+3}\mathcal{F}_{\{0,\pm1\}}(L+2,\normalfont\textbf{p}', s')=\mathcal{F}_{\{0,\pm1/2\}}(L+2,\normalfont\textbf{p}', s')$$ and $$1/4^{L+6}\mathcal{F}_{\{\pm1\}}(L+5,\normalfont\textbf{p}'')=\mathcal{F}_{\{\pm1/4\}}(L+5,\normalfont\textbf{p}'').$$

First, let us show that any network from  $\mathcal{F}_\mathcal{A}(L,\normalfont\textbf{p}, s)$ with weights in $\mathcal{A}=\{0,\pm \frac{1}{2}, \pm 1, 2\}$  can be implemented by a network from $\mathcal{F}_{\{0,\pm\frac{1}{2}\}}(L+2,\normalfont\textbf{p}^\prime, s^\prime)$ with \begin{equation}\label{p1}|\normalfont{\textbf{p}}^\prime|_\infty=4|\normalfont{\textbf{p}}|_\infty\end{equation} and with $s^\prime\leq16s+20(d+1)$. Indeed, using only the weights $0$ and $1/2$ and $2$ hidden layers we can compute the vector 
$$(1,\textbf{x})\mapsto(\smash{\underbrace{1, ... ,1}_{\textrm{4}}, \underbrace{x_1, ... ,x_1}_{\textrm{4}}, ... , \underbrace{x_d, ... ,x_d}_{\textrm{4}} }\vphantom{1}).$$
\vskip 0.2in 
\noindent This computation requires $20(d+1)$ nonzero weights. As each of $0, \pm\frac{1}{2}x, \pm x$ and $2x, x\in\mathbb{R},$ can be represented in the form $\Sigma_{i=1}^4w_ix$ for some $w_i\in\{0, \pm 1/2\}$, $i=1,...,4,$ the desired inclusion \begin{equation}\label{inc1}\mathcal{F}_\mathcal{A}(L,\normalfont\textbf{p}, s)\subset\mathcal{F}_{\{0,\pm1/2\}}(L+2,\normalfont\textbf{p}^\prime, 16s+20(d+1))\end{equation} follows.

Let us now show that networks from  $\mathcal{F}_{\{0,\pm1/2\}}(L+2,\normalfont\textbf{p}^\prime)$ can be implemented by binary weighted networks from $\mathcal{F}_{\{\pm1/4\}}(L+5,\normalfont\textbf{p}^{\prime\prime})$ with \begin{equation}\label{p2}|\normalfont{\textbf{p}}^{\prime\prime}|_\infty=8|\normalfont{\textbf{p}^\prime}|_\infty.\end{equation} It suffices to show that using the weights $\pm1/4$ and $3$ hidden layers we can compute the vector 
$$(1,\textbf{x})\mapsto(1, 1, x_1, x_1, ..., x_d, x_d).$$ The inclusion 
\begin{equation}\label{inc2}
\mathcal{F}_{\{0,\pm1/2\}}(L+2,\normalfont\textbf{p}^\prime)\subset\mathcal{F}_{\{\pm1/4\}}(L+5,\normalfont\textbf{p}^{\prime\prime})
\end{equation}
would then follow as each of $0, \pm\frac{1}{2}x$ can be represented in the form $w_1x+w_2x$ for some $w_1,w_2\in\{\pm 1/4\}$.
Denote $$y_0:=\frac{1}{4}(1+\Sigma_{i=1}^dx_i)\quad \textrm{and} \quad y_k:=\frac{1}{4}(1-x_k+\Sigma_{i\neq k}x_i),$$ $k=1,...,d.$
Then, using the weights $\pm1/4$ and one hidden layer, we can compute the vector 
$$(1,\textbf{x})\mapsto(\smash{\underbrace{y_0, ... ,y_0 }_{\textrm{8}}, \underbrace{y_1, ... ,y_1 }_{\textrm{8}}, ... , \underbrace{y_d, ... ,y_d}_{\textrm{8}} }\vphantom{1}):=\textbf{y}.$$
\vskip 0.2in 
\noindent Note that as $\textbf{x}=(x_1, ... , x_d)\in[0,1]^d,$ then $y_i\geq0,$ and, therefore, the action of the ReLU activation function does not change the value of $y_i,i=0,...,d.$ Also, since each coordinate of the output vector $\textbf{y}$ is repeated $8$ times, then, in the following layer, if needed, each of them can be eliminated by multiplying $4$ equal coordinates by $1/4$ and the other $4$ coordinates by $-1/4$. Hence, as 
$$\frac{1}{4}\cdot8y_0=2y_0\quad \textrm{and}\quad \frac{1}{4}\cdot8y_0-\frac{1}{4}\cdot8y_k=x_k, \quad k=1,...,d,$$ then, using the weights $\pm1/4,$ in the next hidden layer we can compute the vector 
$$\textbf{y}\mapsto(\smash{\underbrace{2y_0, ... ,2y_0 }_{\textrm{8}}, \underbrace{x_1, ... ,x_1 }_{\textrm{4}}, ... , \underbrace{x_d, ... ,x_d}_{\textrm{4}} }\vphantom{1}).$$
\vskip 0.2in 
\noindent
Finally, as $$\frac{1}{4}(16y_0-4\sum_{i=1}^dx_i)=1,$$ then, using the weights $\pm1/4$ and one more hidden layer, we can compute 
$$(\smash{\underbrace{2y_0, ... ,2y_0 }_{\textrm{8}}, \underbrace{x_1, ... ,x_1 }_{\textrm{4}}, ... , \underbrace{x_d, ... ,x_d}_{\textrm{4}} }\vphantom{1})\mapsto (1, 1, x_1, x_1, ..., x_d, x_d)$$\vskip 0.2in 
\noindent and the inclusion \eqref{inc2} follows.

Thus, Theorem \ref{main} now follows combining \eqref{p1}-\eqref{inc2} and using the result of Theorem \ref{prev}.
\newline
\textit{Proof of Theorem \ref{reg}} From \cite{SH}, Lemma 4, it follows that the prediction error of the empirical risk minimizer $\hat{f}_n$ defined in \eqref{erm} is bounded by 
\begin{equation}\label{Oracle} R(\hat{f}_n,f_0)\leq4\bigg[\inf\limits_{f\in\mathcal{F}_n}\EX[\normalfont(f(\textbf{X})-f_0(\textbf{X}))^2]+F^2\frac{18\log_2(\textrm{card}(\mathcal{F}_n))+72}{n}\bigg].\end{equation}

Assume that $f_0\in\mathcal{C}^\beta_d(K)$ with $F\geq\max(K,1)$.
We apply Theorem \ref{main} with $m=m_n=\lceil \log_2n\rceil$ and with $N=N_n=\lceil (\beta+1)^d\lor(K+1)e^d\lor n^{\frac{d}{2\beta+d}}\rceil,$ to get the existence of a network $f\in\mathcal{F}_n:=1/2^{L_n+3}\mathcal{{F}}_{\{0,\pm1\}}({L}_n+2,\normalfont{\textbf{p}}_n,{s}_n)$ such that
\begin{equation}\label{ineq}
\|f-f_0\|^2_{L^\infty[0,1]^d}\leq cn^{\frac{-2\beta}{2\beta+d}},
\end{equation}
where ${L}_n,|\normalfont{\textbf{p}}_n|_\infty$ and ${s}_n$ are defined by \eqref{ln}--\eqref{sn} and $c=c(\beta, d, F)$ is some constant.

Let us now bound the cardinality of the space $\mathcal{F}_n$. The total number of weights in the networks from $\mathcal{F}_n$ is at most $({L}_n+1)|\normalfont{\textbf{p}}_n|_\infty^2$, from which there are at most $s_n$ weights that are equal to $\pm1$ and the remaining weights are equal to $0$. Therefore,
\begin{equation}\label{card}
\textrm{card}(\mathcal{F}_n)\leq\sum\limits_{s\leq{s}_n}\bigg(2({L}_n+1)|\normalfont{\textbf{p}}_n|_\infty^2\bigg)^s\leq\bigg(2({L}_n+1)|\normalfont{\textbf{p}}_n|_\infty^2\bigg)^{{s}_n+1}.\end{equation}
Combining the inequalities \eqref{Oracle}, \eqref{ineq} and \eqref{card} and taking into account \eqref{ln}, \eqref{pn} and \eqref{sn}, we get the desired bound  \eqref{err}.

\end{document}